\documentclass{article}

\usepackage{authblk}

\usepackage[final]{neurips_2024}

\usepackage{times}
\usepackage{soul}
\usepackage{url}
\usepackage[utf8]{inputenc} % allow utf-8 input
\usepackage[T1]{fontenc}    % use 8-bit T1 fonts
\usepackage{hyperref}       % hyperlinks
\usepackage{url}            % simple URL typesetting
\usepackage{booktabs}       % professional-quality tables
\usepackage{amsfonts}       % blackboard math symbols
\usepackage{nicefrac}       % compact symbols for 1/2, etc.
\usepackage{microtype}      % microtypography
\usepackage{xcolor}         % colors
\usepackage[small]{caption}
\usepackage{graphicx}
\usepackage{amsmath}
\usepackage{subfigure}
\usepackage{amsthm}
\usepackage{amssymb,amsfonts}
\usepackage{algpseudocode}
\usepackage[linesnumbered,ruled,vlined]{algorithm2e}
\usepackage{changepage}
\usepackage{subcaption}
\usepackage{multirow}

\newtheorem{theorem}{Theorem}
\title{Machine Unlearning with Minimal Gradient Dependence for High Unlearning Ratios}

\author{%
Tao Huang$^{1*}$ \quad Ziyang Che$^{1*}$ \quad Jiayang Meng$^2$ \quad Qingyu Huang$^1$\\ 
Xu Yang$^1$ \quad Xun Yi$^3$ \quad Ibrahim Khalil$^3$ \\
$^1$ School of Computer Science and Big Data, Minjiang University\\
$^2$ School of Information, Renmin University of China\\
$^3$ RMIT University\\
\texttt{\{huang-tao,ziyangchen,huangqingyu, xuyang\}@mju.edu.cn}\\
\texttt{\{mengjiayang\}@ruc.edu.cn}\\
\texttt{\{Xun Yi,Ibrahim Khalil\}@rmit.edu.au}
}

\begin{document}

\maketitle

\begin{abstract}
In the context of machine unlearning, the primary challenge lies in effectively removing traces of private data from trained models while maintaining model performance and security against privacy attacks like membership inference attacks. Traditional gradient-based unlearning methods often rely on extensive historical gradients, which becomes impractical with high unlearning ratios and may reduce the effectiveness of unlearning. Addressing these limitations, we introduce Mini-Unlearning, a novel approach that capitalizes on a critical observation: unlearned parameters correlate with retrained parameters through contraction mapping. Our method, Mini-Unlearning, utilizes a minimal subset of historical gradients and leverages this contraction mapping to facilitate scalable, efficient unlearning. This lightweight, scalable method significantly enhances model accuracy and strengthens resistance to membership inference attacks. Our experiments demonstrate that Mini-Unlearning not only works under higher unlearning ratios but also outperforms existing techniques in both accuracy and security, offering a promising solution for applications requiring robust unlearning capabilities.
\end{abstract}

\section{Introduction}
\label{sec:introduction}
The widespread adoption of machine learning in critical sectors such as healthcare, finance, and autonomous systems has highlighted the necessity for models that are both adaptable and secure. Traditional machine learning models are designed to continuously learn and retain information. Machine unlearning, as introduced in the literature \cite{b1,b2,b3,b4}, addresses these concerns by enabling models to selectively forget parts of their training data or adjust to changes in data concepts, thus preserving their relevance, accuracy, and confidentiality. A notable approach within machine unlearning is the use of gradient-based techniques \cite{b9,b12,b13,b18,b19,b22,b26,b29,b30}. These methods employ the gradients of the loss function relative to the model parameters to orchestrate the unlearning process. The advantage of gradient-based methods lies in their precise control over the elimination of specific knowledge fragments within a model, allowing for targeted modifications.

In practice, challenges arise when multiple users simultaneously request data removal. In such scenarios, the proportion of data designated for removal—the unlearned data—comprises a significant fraction of the total training data, resulting in a high unlearning ratio. Although gradient-based unlearning methods are effective, they struggle in situations with high unlearning ratios due to their reliance on extensive archives of historical gradients. The reliance often results in unlearned parameters—obtained by erasing gradient information from original parameters—demonstrating reduced test accuracy. Moreover, historical gradients retain sensitive information concerning the unlearned data, making the unlearned parameters derived from them susceptible to privacy attack \cite{b10,b27,b31,b32}.  Additionally, these methods often necessitate further retraining \cite{b19,b22,b26}, which hinders their ability to be implemented in parallel. 

The unlearned parameters obtained via retraining own the highest test accuracy. In our paper, we find the unlearned parameters exhibit a correlation with the retrained parameters via contraction mapping that facilitates the approximation of unlearned parameters with minimal gradient reliance. Leveraging this finding, we introduce 'Mini-Unlearning,' a method that utilizes only a small subset of historical gradients. This method mitigates the adverse effects of discarding excessive historical gradients on the unlearned model's test accuracy and enhances privacy resilience under high unlearning ratios. Moreover, Mini-Unlearning does not require further retraining and can be executed in parallel, provided sufficient computational resources are available. Empirical evaluations on standard datasets verify that Mini-Unlearning effectively maintains accuracy and enhances resistance to membership inference attacks. In summary, our contributions include: 
\begin{itemize}
    \item Our theoretical analysis reveals that the unlearned parameters exhibit a correlation with the retrained parameters via contraction mapping and elucidates the impact of gradients in later training epochs, substantiating its efficacy and parallel capacity.  
    \item Leveraging this critical finding, we introduce Mini-Unlearning, an efficient unlearning method that is effective when the unlearning ratios are large and can be implemented in parallel for the absence of further retraining operations.
    \item We provide enough empirical evidence of superior performances and defensive capabilities of the post-unlearning model achieved through Mini-Unlearning even when facing large unlearning ratios.
\end{itemize}

\section{Related Work}

One way to achieve unlearning goals is by perturbing or masking gradients. \cite{b9} develops an unlearning method based on adaptive query release, which involves efficient gradient recalculations to quickly adjust models when data needs to be forgotten. \cite{b13} presents an unlearning approach called 'Forsaken'. This method employs a mask gradient generator that continuously adjusts gradients to facilitate the unlearning process. \cite{b29} proposes an unlearning method that utilizes a variant of the conditional independence coefficient to identify relevant subsets of model parameters that most influence the data to be forgotten. This approach allows for partial model updates rather than full retraining. \cite{b12} introduces a gradient-based unlearning algorithm that performs gradient descent updates combined with Gaussian noise to ensure statistical indistinguishability from models retrained without the deleted data. \cite{b18} presents a method for approximate data deletion. The main method introduced is the "projective residual update" (PRU), which is gradient-based and reduces the computational cost of data deletion to linear in terms of the data dimension, independent of the dataset size.

To avoid performance degradation caused by gradient modification, machine unlearning based on original parameters or gradients is proposed. \cite{b19} introduces DeltaGrad, an algorithm for rapid retraining through gradient-based methods, which leverages cached training information to update models without retraining from scratch. \cite{b22} introduces certified removal for L2-regularized linear models, which leverages differentiable convex loss functions, and applies a Newton step to adjust model parameters, effectively diminishing the influence of the deleted data point. \cite{b26} introduces Amnesiac Machine Learning which focuses on removing data by leveraging the cached parameters relevant to the deleted data and prevents vulnerability to membership inference attacks while maintaining efficacy.

However, existing methods often require extensive post-processing of numerous historical gradients. Due to specific query types \cite{b9}, intensive manipulation of gradients \cite{b13,b19,b29,b22}, or the challenges in balancing noise addition with gradient updates \cite{b12}, the model's accuracy on the test dataset may be compromised, particularly with large unlearning ratios. Additionally, relying on many historical gradients could undermine the unlearning process, as deleting based on such an extensive set of gradients might not completely eradicate all influences of the unlearned data \cite{b18,b26}. Furthermore, unlearning methods that necessitate additional retraining or much-cached information \cite{b19,b26} lack scalability and cannot be efficiently implemented in parallel. In our paper, we introduce Mini-Unlearning, which utilizes only a small subset of historical gradients to derive the unlearned parameters. Using a smaller subset of historical gradients prevents privacy leakage of derived unlearned parameters, while the contraction mapping ensures the model's test dataset accuracy. Moreover, Mini-Unlearning can be implemented in parallel, significantly enhancing scalability when sufficient computational resources are available.

\section{Notations and Problem Setup}
For the reader's convenience, we collect key notation and background here. 

% In the context of machine learning, the central server collects training samples from various data providers to train a machine learning model denoted as $F(\mathbf{w})$. This model's parameter $\mathbf{w}$ is updated iteratively using the Stochastic Gradient Descent (SGD) method over a fixed number of epochs denoted as $T$. However, data providers, who possess the right to withdraw their data, may decide to remove a subset of samples $\mathbb{D}_U \subseteq \mathbb{D}$ (referred to as "unlearned samples") from the central server's training set. In response, data providers request the server to erase any information related to $\mathbb{D}_U$ from the model's parameter $\mathbf{w}$, thus ensuring their data privacy.

% Formally, the important notations are:
\begin{itemize}
    \item $\mathbb{D}$, $\mathbb{D}_{U}$, $\mathbb{D}_{R}$,$\mathbb{B}_l$, $\Bar{\mathbb{B}}_l$: The full training set $\mathbb{D}$ is composed of two disjoint subsets: $\mathbb{D}_{U}$, representing unlearned samples, and $\mathbb{D}_{R}$, representing retained samples. During the $l$-th iteration of training with a batch $\mathbb{B}_l \subseteq \mathbb{D}$ (with batch size $B$), the set of samples in $\mathbb{B}_l$ from $\mathbb{D}_U$ is denoted as $\Bar{\mathbb{B}}_l$ (namely $\Bar{\mathbb{B}}_l = \mathbb{B}_l \cap \mathbb{D}_U$), with a size of $\Delta B_{l}$.
    \item $F(\mathbf{w})$, $\nabla F^{(l-1,j)}\left(\mathbf{w}_{l-1}\right)$, $\nabla^{2} F^{(l-1,j)}\left(\mathbf{w}_{l-1}\right)$: The machine learning model, $F(\mathbf{w})$, parameterized by $\mathbf{w}$, undergoes updates via SGD with a learning rate of $\eta$. The gradient and Hessian with respect to a training sample $j$ during the $l$-th iteration are $\nabla F^{(l-1,j)}\left(\mathbf{w}_{l-1}\right)$ and $\nabla^{2} F^{(l-1,j)}\left(\mathbf{w}_{l-1}\right)$, respectively.
    \item $\mathbf{w}_T$, $\mathbf{w}_T^{*}$: After the training process's completion, the central server obtains the final model's parameter, $\mathbf{w}_T$. The primary goal is to obtain the "unlearned parameter" denoted as $\mathbf{w}_T^{*}$, which should mimic the behavior of the model trained solely with $\mathbb{D}_{R}$ while not containing any information related to $\mathbb{D}_{U}$.
\end{itemize}

% To achieve the unlearning goal, two conventional methods are considered: retraining the model using $\mathbb{D}_{R}$ or directly approximating $\mathbf{w}_T^{*}$ after $T$ epochs. However, retraining is computationally intensive, and existing approximating methods entail the need to store all historical gradients, which can be storage-intensive. While these methods have demonstrated excellent performance, \textit{\textbf{we argue that many stored gradients may be redundant and a small handful of gradients in the last few epochs is enough for unlearning.}} 

\section{Mini-Unlearning}
\subsection{Main Finding}
\textit{\textbf{We observe that the unlearned parameter $\mathbf{w}_s^{*}$ is primarily influenced by the gradients computed in the last $k$ epochs. many historical gradients may be redundant and a small handful of gradients is enough for unlearning.}} Let's illustrate it step by step.

\paragraph{Starting with the first training epoch.}{The updated model parameter $\mathbf{w}_1$ is computed by the initialized model parameter $\mathbf{w}_0$ minus the batch gradients. That is, $\mathbf{w}_1 = \mathbf{w}_0 -\frac{\eta }{B} \sum_{j \in \mathbb{B}_1} \nabla F^{(0,j)} (\mathbf{w}_0)$. If the data providers want to withdraw the unlearned dataset $\mathbb{D}_{U}$ and quit the training process, the server can first compute $\Bar{\mathbb{B}}_1=\mathbb{B}_1 \cap \mathbb{D}_{U}$. The unlearned parameter $\mathbf{w}_1^{*}$ is $\mathbf{w}_1^{*} = \mathbf{w}_0 -\frac{\eta }{B - \Delta B_1} \sum_{j \in \mathbb{B}_1 / \bar{\mathbb{B}}_{1}} \nabla F^{(0,j)} (\mathbf{w}_0)$. If $\bar{\mathbb{B}}_{1} = \emptyset$, namely SGD takes no unlearned samples to compute gradients, then $\Delta B = 0$ and $\mathbf{w}_1^{*} = \mathbf{w}_1$. Otherwise, we have

\begin{equation}
\small
    \begin{split}
        \Delta \mathbf{w}_1 = \mathbf{w}_1^{*} - \mathbf{w}_1 =\frac{\eta}{B} \left( \frac{B}{B-\Delta B_1}  \sum_{j \in \bar{\mathbb{B}}_{1}} \nabla F^{(0,j)}\left ( \mathbf{w}_{0} \right )\right)-\frac{\eta}{B} \left(\frac{\Delta B_1}{B-\Delta B_1}   \sum_{j \in \mathbb{B}_{1}} \nabla F^{(0,j)}\left ( \mathbf{w}_{0} \right )\right)
    \end{split}
    \label{eq1}
\end{equation}

In other words, instead of retraining, the server can compute $\Delta \mathbf{w}_1$ via $\sum_{j \in \bar{\mathbb{B}}_{1}} \nabla F^{(0,j)}\left ( \mathbf{w}_{0} \right )$. Then the server obtains the unlearned parameter $\mathbf{w}_1^* = \mathbf{w}_1 +\Delta \mathbf{w}_1$. $\sum_{j \in \bar{\mathbb{B}}_{1}} \nabla F^{(0,j)}\left ( \mathbf{w}_{0} \right )$ can be obtained efficiently.}

\paragraph{Moving to the general case.}{What if the data provider wants to withdraw the unlearned dataset $\mathbb{D}_{U}$ and quits the training process after the $s$-th epoch? The server aims to compute $\Delta \mathbf{w}_s$ which satisfies $\mathbf{w}_s^* = \mathbf{w}_s +\Delta \mathbf{w}_s$. Actually, $\mathbf{w}_s^*$ is calculated via Eq.(\ref{eq2}):

\begin{equation}
    \begin{split}
        \mathbf{w}_s^{*} = \mathbf{w}_{s-1}^* -\frac{\eta }{B - \Delta B_s} \sum_{j \in \mathbb{B}_s / \bar{\mathbb{B}}_{s}} \nabla F^{(s-1,j)} (\mathbf{w}_{s-1}^*)
    \end{split}
    \label{eq2}
\end{equation}

$\mathbf{w}_s$ is calculated by: 

\begin{equation}
    \begin{split}
        \mathbf{w}_s = \mathbf{w}_{s-1} -\frac{\eta }{B} \sum_{j \in \mathbb{B}_s} \nabla F^{(s-1,j)} (\mathbf{w}_{s-1})
    \end{split}
    \label{eq3}
\end{equation}

The difference $\Delta \mathbf{w}_s$ between the parameter evaluated on $\mathbb{D}_{R}$ and $\mathbb{D}$ is:
\begin{equation}
\small
    \begin{split}
        &\Delta \mathbf{w}_s= \mathbf{w}_{s}^{*} -\mathbf{w}_s  = \Delta \mathbf{w}_{s-1} - \frac{\eta }{B-\Delta B_s} \sum_{j\in \mathbb{B}_{s}/\bar{\mathbb{B}}_{s}} \nabla F^{(s-1,j)}\left ( \mathbf{w}_{s-1}^{*} \right )\\
        &\quad + \frac{\eta }{B} \left( \sum_{j\in \mathbb{B}_{s}/\bar{\mathbb{B}}_{s}} \nabla F^{(s-1,j)}\left ( \mathbf{w}_{s-1} \right ) + \sum_{j\in \bar{\mathbb{B}}_{s}} \nabla F^{(s-1,j)}\left ( \mathbf{w}_{s-1} \right )\right )
    \end{split}
    \label{eq4}
\end{equation}

Using the Cauchy mean-value theorem, we have:

\begin{equation}
\small
    \begin{split}
        \sum_{j\in \mathbb{B}_{s}/\bar{\mathbb{B}}_{s}} \nabla F^{(s-1,j)}\left ( \mathbf{w}_{s-1}^{*} \right )=\sum_{j\in \mathbb{B}_{s}/\bar{\mathbb{B}}_{s}} \left ( \nabla F^{(s-1,j)}\left ( \mathbf{w}_{s-1} \right ) + \nabla^{2} F^{(s-1,j)}\left ( \mathbf{w}_{s-1} \right ) \cdot \Delta \mathbf{w}_{s-1} \right )
    \end{split}
    \label{eq5}
\end{equation}

Substituting Eq.(\ref{eq5}) into equation Eq.(\ref{eq4}) and denoting $\gamma_s = -\frac{\Delta B_s}{B}$, we can get:

\begin{equation}
\small
    \begin{split}
        &\Delta \mathbf{w}_s =\frac{\eta }{B} \left ( \frac{\gamma_s}{1+\gamma_s}  \right )  \sum_{j\in \mathbb{B}_{s}/\bar{\mathbb{B}}_{s}} \nabla F^{(s-1,j)}\left ( \mathbf{w}_{s-1} \right )-\frac{\eta }{B} \sum_{j\in \bar{\mathbb{B}}_{s}} \nabla F^{(s-1,j)}\left ( \mathbf{w}_{s-1} \right ) \\
        &\quad +  \underset{H(\mathbb{B}_{s},\bar{\mathbb{B}}_{s},\mathbf{w}_{s-1})}{\underbrace{\left ( I - \frac{\eta }{B-\Delta B_s} \sum_{j\in \mathbb{B}_{s}/\bar{\mathbb{B}}_{s}} \nabla^{2} F^{(s-1,j)}\left ( \mathbf{w}_{s-1} \right )  \right )}} \cdot \Delta \mathbf{w}_{s-1}         
    \end{split}
    \label{eq6}
\end{equation}
}

\paragraph{Key finding.}{Assuming that $F$ is $L$-smooth and $\mu$-strongly convex, the maximum eigenvalue of the symmetric positive-defined matrix $H(\mathbb{B}_{s},\bar{\mathbb{B}}_{s},\mathbf{w}_{s-1})$ is $1 - \frac{\eta }{B-\Delta B_s} \cdot \left ( B-\Delta B_s \right) \cdot \mu = 1 - \eta \mu < 1$. Similarly, the minimum eigenvalue is $1 - \eta L < 1$. So the symmetric matrix $H(\mathbb{B}_{s},\bar{\mathbb{B}}_{s},\mathbf{w}_{s-1})$ is \textit{\textbf{a contracting mapping.}} This means $\Delta \mathbf{w}_s$ is \textit{\textbf{mainly influenced by}} $\Delta \mathbf{w}_{s-1},...,\Delta \mathbf{w}_{s-k}$ where $k$ is a small number while $\Delta \mathbf{w}_{s-k-1},..,\Delta \mathbf{w}_{1}$ have negligible impacts on $\Delta \mathbf{w}_s$. Therefore, the gradients and Hessians in the last $k$ epochs are sufficient to compute $\Delta \mathbf{w}_s$. The gradients and Hessians in the earlier epochs are redundant. Motivated by L-BFGS \cite{b28}, the Hessians in $H(\mathbb{B}_{s},\bar{\mathbb{B}}_{s},\mathbf{w}_{s-1})$ can be avoided to calculate and store them. We present the technique to calculate $H(\mathbb{B}_{s},\bar{\mathbb{B}}_{s},\mathbf{w}_{s-1}) \cdot \Delta \mathbf{w}_s$ in section \ref{subsec4.2} and Appendix \ref{Appendix B}.

For simplicity, we define $G\left ( \mathbb{B}_{l}, \bar{\mathbb{B}}_{l}, \mathbf{w}_{l-1}   \right )$ (or $G(l)$) and $H\left ( \mathbb{B}_{l}, \bar{\mathbb{B}}_{l}, \mathbf{w}_{l-1}   \right )$ (or $H(l)$) as follows:

\begin{small}
\begin{equation}
\begin{split}
&G\left ( \mathbb{B}_{l}, \bar{\mathbb{B}}_{l}, \mathbf{w}_{l-1}  \right ) =G\left ( l   \right ):=\frac{\eta}{B} \left ( \frac{\gamma _{l}}{1+\gamma _{l}}  \sum_{j \in \mathbb{B}_{l}/\bar{\mathbb{B}}_{l}} \nabla F^{(l-1,j)}\left ( \mathbf{w}_{l-1} \right ) + \sum_{j \in \bar{\mathbb{B}}_{l}} \nabla F^{(l-1,j)}\left ( \mathbf{w}_{l-1} \right ) \right ) \\
&\quad =\frac{\eta}{B} \underset{\textit{the unlearned samples}}{\underbrace{\left( \frac{B}{B-\Delta B_l}  \sum_{j \in \bar{\mathbb{B}}_{l}} \nabla F^{(l-1,j)}\left ( \mathbf{w}_{l-1} \right )\right)}} -\frac{\eta}{B} \underset{\textit{the batch samples}}{\underbrace{\left(\frac{\Delta B_l}{B-\Delta B_l}   \sum_{j \in \mathbb{B}_{l}} \nabla F^{(l-1,j)}\left ( \mathbf{w}_{l-1} \right )\right)}}  
\end{split}
\label{eq7}
\end{equation}
\end{small}

\begin{equation}
\small
\begin{split}
 H\left ( \mathbb{B}_{l}, \bar{\mathbb{B}}_{l}, \mathbf{w}_{l-1}   \right ) = H\left ( l   \right ) := I - \frac{\eta}{B-\Delta B_{l}} \sum_{j \in \mathbb{B}_{l} /\bar{\mathbb{B}}_{l}} \nabla^{2} F^{(l-1,j)}\left ( \mathbf{w}_{l-1} \right ) 
\end{split}
\label{eq8}
\end{equation}

Substituting Eq.(\ref{eq7}) and Eq.(\ref{eq8}) into Eq.(\ref{eq6}) and pushing forward $k$ rounds, we could infer $\Delta \mathbf{w}_s$ from $\Delta \mathbf{w}_{s-k}$ via Eq.(\ref{eq9}). The derivation of Eq.(\ref{eq9}) could be referred to in Appendix \ref{Appendix A}. 

\begin{equation}
    \begin{split}
    \Delta \mathbf{w}_s = &G\left (s  \right )+ \underset{\text{A}}{\underbrace{\sum_{j=2}^{k} \left ( \prod_{i=1}^{j-1} H\left ( s-i+1  \right ) \right ) \cdot G\left ( s-j+1   \right )}} + \underset{\text{B}}{\underbrace{\left ( \prod_{i=1}^{k} H\left ( s-i+1 \right ) \right ) \cdot \Delta \mathbf{w}_{s-k}}}
    \end{split} 
    \label{eq9}
\end{equation}

Since $H\left ( s-i+1 \right )$ is a contracting mapping, $\Delta \mathbf{w}_s$ can be approximated by setting $\Delta \mathbf{w}_{s-k} = \mathbf{0}$. The approximation is Eq.(\ref{eq10}). The approximation error is displayed in \textbf{Theorem \ref{theorem1}} whose proof is presented in Appendix \ref{Appendix A}.  

\begin{equation}
\small
    \Delta \mathbf{w}_s \approx G\left (s  \right )+ \sum_{j=2}^{k} \left ( \prod_{i=1}^{j-1} H\left ( s-i+1  \right ) \right ) \cdot G\left ( s-j+1   \right )
\label{eq10}
\end{equation}

\begin{theorem}
Suppose $F(\mathbf{w})$ is $\mu$-strongly convex and $L$-smoothness, the approximation error via Eq.(\ref{eq10}) is $o\left(r^k\right)$ where $r = \text{max}\{ \|1-\eta \cdot  \mu \|,  \|1-\eta \cdot  L \|\} \in (0,1)$.
\label{theorem1}
\end{theorem}}

\textbf{Enhanced Explanations. }From Eq.(\ref{eq9}), part B suggests that $\Delta \mathbf{w}_{s-k}$ exerts minimal influence on $\Delta \mathbf{w}_s$, and part A indicates that maintaining the products $\left ( \prod_{i=1}^{j-1} H\left ( s-i+1  \right ) \right ) \cdot G\left ( s-j+1   \right )$ suffices, ensuring that the original gradients $G(s-j+1)$ are not exposed to adversaries. Moreover, the hyper-parameter $k$ in part A specifies the extent of historical gradients utilized to compute $\Delta \mathbf{w}_s$. A smaller $k$ results in $\Delta \mathbf{w}_s$ incorporating less private gradient information, thereby enhancing the robustness of $\mathbf{w}_s^{*}$. 

\subsection{Implementations of Mini-Unlearning}\label{subsec4.2}
\begin{algorithm}[t] 
\small
	\caption{Mini-Unlearning} 
	\label{alg1}
            \KwIn{the full training set $\mathbb{D}$; the model $F(\mathbf{w})$; the initial model's parameter $\mathbf{w}_0$; the batch size $B$ of SGD; the unlearned set $\mathbb{D}_{U}$; the training epoch number $T$.}
            \KwOut{the unlearned parameter $\mathbf{w}_{T}^{*}$.}
            Select $\{ \mathbf{u}_1,...,\mathbf{u}_p \} \in \mathbb{R}^p$ \\
            $\Delta H = []$, $\Delta G = []$ \\
            $\Delta H[0].append([\mathbf{u}_1,...,\mathbf{u}_p])$ \\
            \For{$i = 1; i++; i \le T$}{   
                $\mathbb{B}_{i} = \text{Sample} \left( \mathbb{D} \right)$ \\
                Compute $\sum_{j \in \mathbb{B}_{i}} \nabla F^{(i-1,j)}\left ( \mathbf{w}_{i-1} \right )$ \\
                \tcp*[h]{Calculate and Store Information in Last $k$ Epochs} \\
                \If{$i \geq T-k$}{
                    $n=0$ \\
                    Compute $\sum_{j \in \bar{\mathbb{B}}_{i}} \nabla F^{(i-1,j)}\left ( \mathbf{w}_{i-1} \right )$ \\
                    Compute $G(i)$ by Eq.(\ref{eq7}) \\
                    $\Delta G.append(G(i))$ \\
                    $\Delta H[n+1][q] \leftarrow H(i) \cdot \Delta H[n][q]$ for $q=1,2,...,p$ \\
                    $n++$}}
            \tcp*[h]{Calculate $\Delta \mathbf{w}_T$} \\
            Set $\Delta \mathbf{w}_T = 0$ \\
            \For{$r=1; r<k-1; r++$}{
                $\Delta \mathbf{w}_T \leftarrow \Delta \mathbf{w}_T +\sum_{q=1}^{p} \left( \Delta H[r][q] \cdot \left(\Delta H[0][q]\right)^T \cdot \Delta G(r+1) \right)$ \\
            }
            $\begin{small}
                \Delta \mathbf{w}_T \leftarrow \Delta \mathbf{w}_T + \Delta G(k-1)
            \end{small}$ \\
            \Return{$\mathbf{w}_T^* = \mathbf{w}_T + \Delta \mathbf{w}_T$}  
\end{algorithm}

\begin{algorithm}[htb]
\small
	\caption{Init Training} 
	\label{alg3} 
            \KwIn{the full training set $\mathbb{D}$; the model $F(\mathbf{w})$; the initial model's parameter $\mathbf{w}_0$; the batch size $B$ of SGD; Hyperparameter $k$; the unlearned set $\mathbb{D}_U$; the training epoch number $T$}
            \KwOut{model's parameters $ W = \{ \mathbf{w}_{T-k},...,\mathbf{w}_T \}$; gradients $G(T-k), \cdots ,G(T)$; Sample Index List $InL$.}
            Initialize the model's parameter $\mathbf{w}_0$ \\
            $InL = []$, $ W = []$, $G = []$ \\
            \tcp*[h]{Model Updating Via SGD} \\
            \For{$i = 1; i++; i \leq T;$}{
                $\mathbb{B}_{i} = \text{Sample} \left( \mathbb{D} \right)$ \\ 
                $\mathbf{w}_i \leftarrow \mathbf{w}_{i-1} - \frac{\eta}{B} \sum_{j \in \mathbb{B}_{i}} \nabla F^{(i-1,j)} \left ( \mathbf{w}_{i-1} \right )$ \\
                \tcp*[h]{Calculate and Store Information in Last $k$ Epochs} \\
                \If{$i\geq T-k$}{
                    Obtain $\mathbb{B}_{i} / \Bar{\mathbb{B}}_{i}$ \\
                    Store the indices of all samples in $\mathbb{B}_{i} / \Bar{\mathbb{B}}_{i}$ in $InL$ \\
                    Compute $\sum_{j \in \Bar{\mathbb{B}}_{i}} \nabla F^{(i-1,j)}\left ( \mathbf{w}_{i-1} \right )$ \\
                    Compute $G(i)$ by Eq.(\ref{eq7}) \\
                    $ G.append(G(i))$ \\
                    $ W.append(w_i)$ \\
                }
            }  
            \Return{$W$, $G$, $InL$} 
\end{algorithm}

\begin{algorithm}[tb]
\small
	\caption{Mini-Unlearning in Parallel} 
	\label{alg2} 
            \KwIn{the full training set $\mathbb{D}$; the model $F(\mathbf{w})$; the batch size $B$ of SGD; $W$, $G$, $InL$ obtained from Algorithm \ref{alg3}.} 
            \KwOut{the unlearned model's parameter $\mathbf{w}_{T}^{*}$.}
            Select $\{ \mathbf{u}_1,...,\mathbf{u}_p \} \in \mathbb{R}^p$\\
            $\Delta H = []$\\
            \tcp*[h]{Calculating in Parallel} \\
            \For{$i \in \{ 0,1,...,k-1 \}$}{
                Obtain $\mathbb{B}_{T-k+i} / \Bar{\mathbb{B}}_{T-k+i}$ via $InL$ \\
                Compute $H(T-k+i) \cdot \mathbf{u}_q $ for $q=1,2,...,p$ \\
                $H(T-k+i) = \sum_{q=1}^{p} H(T-k+i) \cdot \mathbf{u}_q \cdot (\mathbf{u}_q)^T$ \\
                $\Delta H.append(H(T-k+i))$      
            } 
            \tcp*[h]{Calculate $\Delta \mathbf{w}_T$} \\
            Set $\Delta \mathbf{w}_{T-k} = 0$\\
            Compute $\Delta \mathbf{w}_T$ by Eq.(\ref{eq10})\\
            \Return{$\mathbf{w}_T^* = \mathbf{w}_T + \Delta \mathbf{w}_T$} 
\end{algorithm}

\paragraph{A small handful of gradients is enough.} {From Eq.(\ref{eq10}), we need $\{G(s-j+1)\}_{j=1}^k$. The smaller $k$, the fewer historical gradients. Back to Eq.(\ref{eq7}), the batch samples could be calculated directly via SGD where no more computation is needed. The unlearned samples could be calculated via PyTorch libraries like Opacus \footnote{https://opacus.ai/} or Backpack \footnote{https://docs.backpack.pt/en/master/index.html}.

\paragraph{Calculating the products of symmetric matrices. \label{para4.2}} {From Eq.(\ref{eq10}), we need the products of a series of symmetric matrices to approximate $\Delta \mathbf{w}_s$. However, the computation cost to compute $\prod_{i=1}^{j-1} H\left ( s-i+1  \right )$ directly is high. Inspired by L-BFGS \cite{b28}, we present an efficient method to compute $\prod_{i=1}^{j-1} H\left ( s-i+1  \right )$ by calculating $H(s-i+1) \cdot \mathbf{u}$ where $\mathbf{u}$ is a constant vector independent of the model parameter $\mathbf{w}_{s-i+1}$. The method is presented in Appendix \ref{Appendix B}.

To calculate $H(l)$, we first choose orthonormal basis $\{ \mathbf{u}_1,...,\mathbf{u}_p \}$ from $p$-dimensional Euclidean space $\mathbb{R}^p$ where $p = \text{dimension}(\mathbf{w})$. Then we compute the series $\{H(l)\cdot \mathbf{u}_1,..,H(l)\cdot \mathbf{u}_p\}$. To recover $H(l)$, we compute $\{H(l)\cdot \mathbf{u}_1 \cdot (\mathbf{u}_1)^T,...,H(l)\cdot \mathbf{u}_p \cdot (\mathbf{u}_p)^T\}$. Subsequently, $\sum_{q=1}^{p} H(l)\cdot \mathbf{u}_q \cdot (\mathbf{u}_q)^T = H(l) \cdot \left( \sum_{q=1}^{p} \mathbf{u}_q \cdot (\mathbf{u}_q)^T \right) = H(l) \cdot I = H(l)$. Moreover, when $\{ \mathbf{u}_1,...,\mathbf{u}_p \}$ are the standard basis, the computation could be more efficient. 

When calculating $H(l) \cdot H(l+1)$, given a constant vector $\mathbf{u}$, it should be noted that $\Bar{\mathbf{u}} = H(l) \cdot \mathbf{u}$ can be regarded as a constant vector independent of $H(l+1)$ and we can calculate $H(l+1) \cdot \Bar{\mathbf{u}} = H(l+1) \cdot H(l) \cdot \mathbf{u}$. Then $H(l) \cdot H(l+1)$ can be recovered. The products $H(l) \cdot H(l+1) \cdot \cdots \cdot H(l+m)$ could be obtained inductively.}

\paragraph{Algorithm details.}{Algorithm~\ref{alg1} presents the procedure of our unlearning method. It deals with the situation where the data provider withdraws the unlearned data $\mathbb{D}_{U}$ after $T$ epochs. The server computes $\mathbf{w}_T^*$, deletes $\mathbb{D}_{U}$ from $\mathbb{D}$, and then goes on training with $\mathbb{D}_{R} = \mathbb{D} / \mathbb{D}_{U}$.}

We also present a more efficient way to implement Mini-Unlearning in parallel for the server configured with enough computation resources. Supposing that the server has $k$ devices. Given constant vectors $\{\mathbf{u}_1,...,\mathbf{u}_p \}$, $H(T-k+i) \cdot \{\mathbf{u}_q\}_{q=1}^{p}$ is calculated on the $i$-th device via the method presented in Appendix \ref{Appendix B}. It should be noted that the Hessian is symmetric. Namely, $H(s) = H(s)^T$. For $H(l)\cdot H(s)$, we get:
\begin{equation}
\small
\begin{split}
   &\sum_{q=1}^{p} \left ( (H(l)\cdot \mathbf{u}_q) \cdot (H(s) \cdot \mathbf{u}_q)^T \right) = \sum_{q=1}^{p} \left ( H(l) \cdot \mathbf{u}_q \cdot (\mathbf{u}_q)^T \cdot H(s) \right)  \\
   &\quad = H(l) \cdot \sum_{q=1}^{p} \left (  \mathbf{u}_q \cdot (\mathbf{u}_q)^T  \right) \cdot H(s) = H(l) \cdot I \cdot H(s) = H(l) \cdot H(s) 
\end{split}
\end{equation}

After all the Hessians and their products are recovered, the server could compute $\Delta \mathbf{w}_s$ directly by Eq.(\ref{eq10}). The two-phrase unlearning method is presented in Algorithms ~\ref{alg2}, together with the init training Algorithm ~\ref{alg3}. Algorithm~\ref{alg3} gives the init training process whose outputs are parameters and gradients in the last $k$ epochs while Algorithm~\ref{alg2} presents the parallel machine unlearning process, which computes $\Delta \mathbf{w}_T$.

\section{Experiments}

\subsection{Experiment Setup}
%and RCV1\cite{b26}.
\paragraph{Datasets and Settings.} In our experiments, three benchmark datasets are employed to evaluate the performances. They are \textit{MNIST}\cite{b23}, \textit{Covtype}\cite{b24}, and \textit{HIGGS}\cite{b25}. We evaluate our method over regularized logistic regression over the three datasets with an L2 norm coefficient of 0.005 and, a fixed learning rate of 0.01. The training algorithm is SGD. $k$ is set to 10. To simulate the elimination of training instances, we manipulate various unlearning ratios, $\gamma$, where $\gamma = \frac{\mathbb{D}_U}{\mathbb{D}}$. Contrary to prior studies where $\gamma$ is maintained at a relatively minuscule value (e.g., $0.005\%$), in our experimental framework, we set $\gamma$ to large values, specifically $5\%$, $10\%$, and $15\%$. This adjustment is made to ascertain the efficacy of Mini-Unlearning in scenarios characterized by higher data removal ratios. All experiments are run over a GPU machine with one Intel(R) Xeon(R) Silver 4210 CPU @ 2.20GHz with 128 GB DRAM and 4 GeForce 4090 RTX GPUs (each GPU has 24 GB DRAM). We implemented Mini-Unlearning with PyTorch 1.3.

\paragraph{Baselines.} Our baselines consist of Traditional Retraining, DeltaGrad \cite{b19}, Certified Data Removal \cite{b22}, and Amnesiac Unlearning \cite{b26}. We store the unlearned models obtained by baselines for future evaluations.

\emph{Traditional Retraining.} Traditional retraining could be seen as the upper bound of the model's performance when facing machine unlearning since it only considers the utility of the model but does not take efficiency into account. This baseline is implemented by removing the unlearned samples and retraining from the beginning. Retraining keeps the same initialization and hyperparameter setting as the original training process.

\emph{DeltaGrad. \cite{b19}} DeltaGrad is demonstrated to achieve state-of-the-art performance when the fraction of unlearned samples is relatively small. We keep the same parameter settings as in Delta. Namely, we set $T_0 = 5, j_0 = 10$ for MNIST and Covtype, and $T_0 =3, j_0 = 300$ for HIGGS.

\emph{Certified Data Removal. \cite{b22}} Certified data removal from the linear logistic regression has two parameters to be set, the regularization parameter and the standard deviation. The regularization parameter has been set as 0.005. We choose the standard deviation $\sigma = 0.01$ where Certified data removal shows the best performance in the original paper.

\emph{Amnesiac Unlearning. \cite{b26}} Amnesiac Unlearning performs unlearning operations when receiving data removal requests. Thus, after finishing training the original model, we perform amnesiac Unlearning directly to obtain the unlearning model. No special parameters need to be mentioned.

\paragraph{Evaluation Metrics.} In assessing the effectiveness of diverse unlearning methodologies, our focus is twofold: firstly, on the performance metrics of models post-unlearning, and secondly, on the resilience of these models against potential security breaches. We appraise the test accuracy of $F(\mathbf{w}_T^*)$ across the test set. To ascertain the thoroughness of unlearned data exclusion from $\mathbb{D}_U$, we execute a membership inference attack (MIA) on $F(\mathbf{w}_T^*)$ where MIA is designed to detect whether a data sample is used for model training or not. Each data sample $d_i \in \mathbb{D}_U$ is scrutinized under this framework. To illustrate that the information about the retained data in $\mathbb{D}_R$ is remaining in $\mathbf{w}_T^*$, we also scrutinize data sample $d_j \in \mathbb{D}_R$ under this framework. An attack model is trained to determine the membership status of each $d_i, d_j$. The efficacy of this approach is measured using precision and recall metrics.

\subsection{Experiment Results}

\begin{table}[]
\caption{Performance of Unlearned Models}
\centering
\small
\begin{tabular}{ccccc}
\toprule[1pt]
% \multicolumn{5}{c}{\textbf{Performance of Unlearned Models}}                                                                                                                               \\ \toprule
\multicolumn{2}{c|}{\textbf{Dataset}}                                                                                  & \multicolumn{1}{c|}{\textbf{MNIST}} & \multicolumn{1}{c|}{\textbf{Covtype}} & \textbf{HIGGS} \\ \toprule[1pt]
\textbf{Unlearning Ratios}     & \multicolumn{1}{c|}{\textbf{Methods}}                                                          & \multicolumn{3}{c}{Test Accuracy $\uparrow$}                                 \\ \toprule[1pt]
\multirow{5}{*}{5\%}  & \multicolumn{1}{c|}{Retraining}                                                       & \multicolumn{1}{c|}{0.85}  & \multicolumn{1}{c|}{0.78}    & 0.7   \\ \cline{2-5} 
                      & \multicolumn{1}{c|}{DeltaGrad}                                                        & \multicolumn{1}{c|}{0.78}  & \multicolumn{1}{c|}{0.65}    & 0.46  \\ \cline{2-5} 
                      & \multicolumn{1}{c|}{\begin{tabular}[c]{@{}c@{}}Certified Data Removal\end{tabular}} & \multicolumn{1}{c|}{0.33}  & \multicolumn{1}{c|}{0.13}    & 0.36  \\ \cline{2-5} 
                      & \multicolumn{1}{c|}{\begin{tabular}[c]{@{}c@{}}Amnesiac Unlearning\end{tabular}}    & \multicolumn{1}{c|}{0.29}  & \multicolumn{1}{c|}{0.21}    & 0.23  \\ \cline{2-5} 
                      & \multicolumn{1}{c|}{Mini-Unlearning}                                                  & \multicolumn{1}{c|}{\textbf{0.82}}  & \multicolumn{1}{c|}{\textbf{0.70}}    & \textbf{0.64}  \\ \midrule
\multirow{5}{*}{10\%} & \multicolumn{1}{c|}{Retraining}                                                       & \multicolumn{1}{c|}{0.82}  & \multicolumn{1}{c|}{0.77}    & 0.67  \\ \cline{2-5} 
                      & \multicolumn{1}{c|}{DeltaGrad}                                                        & \multicolumn{1}{c|}{0.73}  & \multicolumn{1}{c|}{0.47}    & 0.44  \\ \cline{2-5} 
                      & \multicolumn{1}{c|}{\begin{tabular}[c]{@{}c@{}}Certified Data Removal\end{tabular}} & \multicolumn{1}{c|}{0.25}  & \multicolumn{1}{c|}{0.12}    & 0.33  \\ \cline{2-5} 
                      & \multicolumn{1}{c|}{\begin{tabular}[c]{@{}c@{}}Amnesiac Unlearning\end{tabular}}    & \multicolumn{1}{c|}{0.22}  & \multicolumn{1}{c|}{0.21}    & 0.23  \\ \cline{2-5} 
                      & \multicolumn{1}{c|}{Mini-Unlearning}                                                  & \multicolumn{1}{c|}{\textbf{0.79}}  & \multicolumn{1}{c|}{\textbf{0.67}}    & \textbf{0.58}  \\ \midrule
\multirow{5}{*}{15\%} & \multicolumn{1}{c|}{Retraining}                                                       & \multicolumn{1}{c|}{0.79}  & \multicolumn{1}{c|}{0.77}    & 0.66  \\ \cline{2-5} 
                      & \multicolumn{1}{c|}{DeltaGrad}                                                        & \multicolumn{1}{c|}{0.67}  & \multicolumn{1}{c|}{0.44}    & 0.40  \\ \cline{2-5} 
                      & \multicolumn{1}{c|}{\begin{tabular}[c]{@{}c@{}}Certified Data Removal\end{tabular}} & \multicolumn{1}{c|}{0.31}  & \multicolumn{1}{c|}{0.13}    & 0.33  \\ \cline{2-5} 
                      & \multicolumn{1}{c|}{\begin{tabular}[c]{@{}c@{}}Amnesiac Unlearning\end{tabular}}    & \multicolumn{1}{c|}{0.21}  & \multicolumn{1}{c|}{0.20}    & 0.21  \\ \cline{2-5} 
                      & \multicolumn{1}{c|}{Mini-Unlearning}                                                  & \multicolumn{1}{c|}{\textbf{0.74}}  & \multicolumn{1}{c|}{\textbf{0.66}}    & \textbf{0.56}  \\ \bottomrule[1pt]
\end{tabular}
\label{table1}
\end{table}

\begin{table}[]
\caption{Defensive Ability over MIA}
\small
\begin{tabular}{ccclclclclclll}
\toprule[1pt]
% \multicolumn{14}{c}{\textbf{Defensive Ability over MIA}}                                                                                                                                                                                                                                                                                                                                                                                                                                                                                                                                                                                                                                                                                \\ \toprule
\multicolumn{2}{c|}{\multirow{2}{*}{\textbf{Dataset}}}                                                                                                        & \multicolumn{4}{c|}{\textbf{MNIST}}                                                                                                                                                 & \multicolumn{4}{c|}{\textbf{Covtype}}                                                                                                                                               & \multicolumn{4}{c}{\textbf{HIGGS}}                                                                                                                                                 \\ \cline{3-14} 
\multicolumn{2}{c|}{}                                                                                                                                         & \multicolumn{2}{c}{\textit{\begin{tabular}[c]{@{}c@{}}Unlearned \\ Sample\end{tabular}}} & \multicolumn{2}{c|}{\textit{\begin{tabular}[c]{@{}c@{}}Retained\\  Sample\end{tabular}}} & \multicolumn{2}{c}{\textit{\begin{tabular}[c]{@{}c@{}}Unlearned \\ Sample\end{tabular}}} & \multicolumn{2}{c|}{\textit{\begin{tabular}[c]{@{}c@{}}Retained\\  Sample\end{tabular}}} & \multicolumn{2}{c}{\textit{\begin{tabular}[c]{@{}c@{}}Unlearned \\ Sample\end{tabular}}} & \multicolumn{2}{c}{\textit{\begin{tabular}[c]{@{}c@{}}Retained \\ Sample\end{tabular}}} \\ \toprule[1pt]
\textbf{\begin{tabular}[c]{@{}c@{}}Unlearning \\ Ratios\end{tabular}} & \multicolumn{1}{c|}{\textbf{Methods}}                                                 & \multicolumn{4}{c|}{\begin{tabular}[c]{@{}c@{}}Precision $\downarrow$\\ (Recall $\downarrow$)\end{tabular}}                                                                                                   & \multicolumn{4}{c|}{\begin{tabular}[c]{@{}c@{}}Precision $\downarrow$\\ (Recall $\downarrow$)\end{tabular}}                                                                                                   & \multicolumn{4}{c}{\begin{tabular}[c]{@{}c@{}}Precision $\downarrow$\\ (Recall $\downarrow$)\end{tabular}}                                                                                                   \\ \toprule[1pt]
\multirow{5}{*}{5\%}                                                  & \multicolumn{1}{c|}{Retraining}                                                       & \multicolumn{2}{c}{\begin{tabular}[c]{@{}c@{}}0.6995\\ (0.7425)\end{tabular}}            & \multicolumn{2}{c|}{\begin{tabular}[c]{@{}c@{}}0.7554\\ (0.7998)\end{tabular}}           & \multicolumn{2}{c}{\begin{tabular}[c]{@{}c@{}}0.6063\\ (0.6866)\end{tabular}}            & \multicolumn{2}{c|}{\begin{tabular}[c]{@{}c@{}}0.7203\\ (0.7039)\end{tabular}}           & \multicolumn{2}{c}{\begin{tabular}[c]{@{}c@{}}0.5946\\ (0.6044)\end{tabular}}            & \multicolumn{2}{l}{\begin{tabular}[c]{@{}l@{}}0.6944\\ (0.6354)\end{tabular}}           \\ \cline{2-14} 
                                                                      & \multicolumn{1}{c|}{DeltaGrad}                                                        & \multicolumn{2}{c}{\begin{tabular}[c]{@{}c@{}}0.6566\\ (0.7025)\end{tabular}}            & \multicolumn{2}{c|}{\begin{tabular}[c]{@{}c@{}}0.6996\\ (0.7445)\end{tabular}}           & \multicolumn{2}{c}{\begin{tabular}[c]{@{}c@{}}0.5654\\ (0.5916)\end{tabular}}            & \multicolumn{2}{c|}{\begin{tabular}[c]{@{}c@{}}0.6562\\ (0.6459)\end{tabular}}           & \multicolumn{2}{c}{\begin{tabular}[c]{@{}c@{}}0.5665\\ (0.5912)\end{tabular}}            & \multicolumn{2}{l}{\begin{tabular}[c]{@{}l@{}}0.6379\\ (0.6432)\end{tabular}}           \\ \cline{2-14} 
                                                                      & \multicolumn{1}{c|}{\begin{tabular}[c]{@{}c@{}}Certified Data\\ Removal\end{tabular}} & \multicolumn{2}{c}{\begin{tabular}[c]{@{}c@{}}0.5155\\ (0.5033)\end{tabular}}            & \multicolumn{2}{c|}{\begin{tabular}[c]{@{}c@{}}0.5002\\ (0.4879)\end{tabular}}           & \multicolumn{2}{c}{\begin{tabular}[c]{@{}c@{}}0.4899\\ (0.4554)\end{tabular}}            & \multicolumn{2}{c|}{\begin{tabular}[c]{@{}c@{}}0.4807\\ (0.4634)\end{tabular}}           & \multicolumn{2}{c}{\begin{tabular}[c]{@{}c@{}}0.4996\\ (0.4658)\end{tabular}}            & \multicolumn{2}{l}{\begin{tabular}[c]{@{}l@{}}0.4887\\ (0.4706)\end{tabular}}           \\ \cline{2-14} 
                                                                      & \multicolumn{1}{c|}{\begin{tabular}[c]{@{}c@{}}Amnesiac\\ Unlearning\end{tabular}}    & \multicolumn{2}{c}{\begin{tabular}[c]{@{}c@{}}0.4950\\ (0.5051)\end{tabular}}            & \multicolumn{2}{c|}{\begin{tabular}[c]{@{}c@{}}0.4968\\ (0.4996)\end{tabular}}           & \multicolumn{2}{c}{\begin{tabular}[c]{@{}c@{}}0.4783\\ (0.4705)\end{tabular}}            & \multicolumn{2}{c|}{\begin{tabular}[c]{@{}c@{}}0.4799\\ (0.4695)\end{tabular}}           & \multicolumn{2}{c}{\begin{tabular}[c]{@{}c@{}}0.4895\\ (0.4978)\end{tabular}}            & \multicolumn{2}{l}{\begin{tabular}[c]{@{}l@{}}0.4903\\ (0.4901)\end{tabular}}           \\ \cline{2-14} 
                                                                      & \multicolumn{1}{c|}{Mini-Unlearning}                                                  & \multicolumn{2}{c}{\textbf{\begin{tabular}[c]{@{}c@{}}0.5181\\ (0.6503)\end{tabular}}}   & \multicolumn{2}{c|}{\textbf{\begin{tabular}[c]{@{}c@{}}0.6003\\ (0.6979)\end{tabular}}}  & \multicolumn{2}{c}{\textbf{\begin{tabular}[c]{@{}c@{}}0.5166\\ (0.5043)\end{tabular}}}   & \multicolumn{2}{c|}{\textbf{\begin{tabular}[c]{@{}c@{}}0.6033\\ (0.6776)\end{tabular}}}  & \multicolumn{2}{c}{\textbf{\begin{tabular}[c]{@{}c@{}}0.4494\\ (0.5943)\end{tabular}}}   & \multicolumn{2}{l}{\textbf{\begin{tabular}[c]{@{}l@{}}0.5338\\ (0.6019)\end{tabular}}}  \\ \midrule
\multirow{5}{*}{10\%}                                                 & \multicolumn{1}{c|}{Retraining}                                                       & \multicolumn{2}{c}{\begin{tabular}[c]{@{}c@{}}0.6557\\ (0.7184)\end{tabular}}            & \multicolumn{2}{c|}{\begin{tabular}[c]{@{}c@{}}0.7605\\ (0.8018)\end{tabular}}           & \multicolumn{2}{c}{\begin{tabular}[c]{@{}c@{}}0.5759\\ (0.6442)\end{tabular}}            & \multicolumn{2}{c|}{\begin{tabular}[c]{@{}c@{}}0.7263\\ (0.7056)\end{tabular}}           & \multicolumn{2}{c}{\begin{tabular}[c]{@{}c@{}}0.5658\\ (0.5722)\end{tabular}}            & \multicolumn{2}{l}{\begin{tabular}[c]{@{}l@{}}0.6989\\ (0.6395)\end{tabular}}           \\ \cline{2-14} 
                                                                      & \multicolumn{1}{c|}{DeltaGrad}                                                        & \multicolumn{2}{c}{\begin{tabular}[c]{@{}c@{}}0.6018\\ (0.6891)\end{tabular}}            & \multicolumn{2}{c|}{\begin{tabular}[c]{@{}c@{}}0.7017\\ (0.7364)\end{tabular}}           & \multicolumn{2}{c}{\begin{tabular}[c]{@{}c@{}}0.5458\\ (0.5887)\end{tabular}}            & \multicolumn{2}{c|}{\begin{tabular}[c]{@{}c@{}}0.6601\\ (0.6468)\end{tabular}}           & \multicolumn{2}{c}{\begin{tabular}[c]{@{}c@{}}0.5446\\ (0.5794)\end{tabular}}            & \multicolumn{2}{l}{\begin{tabular}[c]{@{}l@{}}0.6401\\ (0.6464)\end{tabular}}           \\ \cline{2-14} 
                                                                      & \multicolumn{1}{c|}{\begin{tabular}[c]{@{}c@{}}Certified Data\\ Removal\end{tabular}} & \multicolumn{2}{c}{\begin{tabular}[c]{@{}c@{}}0.5106\\ (0.5003)\end{tabular}}            & \multicolumn{2}{c|}{\begin{tabular}[c]{@{}c@{}}0.5017\\ (0.4909)\end{tabular}}           & \multicolumn{2}{c}{\begin{tabular}[c]{@{}c@{}}0.5001\\ (0.4982)\end{tabular}}            & \multicolumn{2}{c|}{\begin{tabular}[c]{@{}c@{}}0.4875\\ (0.4601)\end{tabular}}           & \multicolumn{2}{c}{\begin{tabular}[c]{@{}c@{}}0.4901\\ (0.4705)\end{tabular}}            & \multicolumn{2}{l}{\begin{tabular}[c]{@{}l@{}}0.4809\\ (0.4712)\end{tabular}}           \\ \cline{2-14} 
                                                                      & \multicolumn{1}{c|}{\begin{tabular}[c]{@{}c@{}}Amnesiac\\ Unlearning\end{tabular}}    & \multicolumn{2}{c}{\begin{tabular}[c]{@{}c@{}}0.4879\\ (0.4998)\end{tabular}}            & \multicolumn{2}{c|}{\begin{tabular}[c]{@{}c@{}}0.4901\\ (0.4903)\end{tabular}}           & \multicolumn{2}{c}{\begin{tabular}[c]{@{}c@{}}0.4894\\ (0.4768)\end{tabular}}            & \multicolumn{2}{c|}{\begin{tabular}[c]{@{}c@{}}0.4707\\ (0.4705)\end{tabular}}           & \multicolumn{2}{c}{\begin{tabular}[c]{@{}c@{}}0.4804\\ (0.4866)\end{tabular}}            & \multicolumn{2}{l}{\begin{tabular}[c]{@{}l@{}}0.4912\\ (0.4887)\end{tabular}}           \\ \cline{2-14} 
                                                                      & \multicolumn{1}{c|}{Mini-Unlearning}                                                  & \multicolumn{2}{c}{\textbf{\begin{tabular}[c]{@{}c@{}}0.4964\\ (0.6324)\end{tabular}}}   & \multicolumn{2}{c|}{\textbf{\begin{tabular}[c]{@{}c@{}}0.6099\\ (0.7077)\end{tabular}}}  & \multicolumn{2}{c}{\textbf{\begin{tabular}[c]{@{}c@{}}0.4213\\ (0.4299)\end{tabular}}}   & \multicolumn{2}{c|}{\textbf{\begin{tabular}[c]{@{}c@{}}0.6064\\ (0.6798)\end{tabular}}}  & \multicolumn{2}{c}{\textbf{\begin{tabular}[c]{@{}c@{}}0.4482\\ (0.5532)\end{tabular}}}   & \multicolumn{2}{l}{\textbf{\begin{tabular}[c]{@{}l@{}}0.5355\\ (0.6008)\end{tabular}}}  \\ \midrule
\multirow{5}{*}{15\%}                                                 & \multicolumn{1}{c|}{Retraining}                                                       & \multicolumn{2}{c}{\begin{tabular}[c]{@{}c@{}}0.6253\\ (0.6891)\end{tabular}}            & \multicolumn{2}{c|}{\begin{tabular}[c]{@{}c@{}}0.7663\\ (0.8057)\end{tabular}}           & \multicolumn{2}{c}{\begin{tabular}[c]{@{}c@{}}0.5356\\ (0.6089)\end{tabular}}            & \multicolumn{2}{c|}{\begin{tabular}[c]{@{}c@{}}0.7263\\ (0.7056)\end{tabular}}           & \multicolumn{2}{c}{\begin{tabular}[c]{@{}c@{}}0.5439\\ (0.5578)\end{tabular}}            & \multicolumn{2}{l}{\begin{tabular}[c]{@{}l@{}}0.6904\\ (0.6415)\end{tabular}}           \\ \cline{2-14} 
                                                                      & \multicolumn{1}{c|}{DeltaGrad}                                                        & \multicolumn{2}{c}{\begin{tabular}[c]{@{}c@{}}0.5865\\ (0.6324)\end{tabular}}            & \multicolumn{2}{c|}{\begin{tabular}[c]{@{}c@{}}0.7009\\ (0.7419)\end{tabular}}           & \multicolumn{2}{c}{\begin{tabular}[c]{@{}c@{}}0.5146\\ (0.5579)\end{tabular}}            & \multicolumn{2}{c|}{\begin{tabular}[c]{@{}c@{}}0.6601\\ (0.6468)\end{tabular}}           & \multicolumn{2}{c}{\begin{tabular}[c]{@{}c@{}}0.5169\\ (0.5368)\end{tabular}}            & \multicolumn{2}{l}{\begin{tabular}[c]{@{}l@{}}0.6422\\ (0.6477)\end{tabular}}           \\ \cline{2-14} 
                                                                      & \multicolumn{1}{c|}{\begin{tabular}[c]{@{}c@{}}Certified Data\\ Removal\end{tabular}} & \multicolumn{2}{c}{\begin{tabular}[c]{@{}c@{}}0.5005\\ (0.5021)\end{tabular}}            & \multicolumn{2}{c|}{\begin{tabular}[c]{@{}c@{}}0.5022\\ (0.4867)\end{tabular}}           & \multicolumn{2}{c}{\begin{tabular}[c]{@{}c@{}}0.5013\\ (0.4992)\end{tabular}}            & \multicolumn{2}{c|}{\begin{tabular}[c]{@{}c@{}}0.4875\\ (0.4601)\end{tabular}}           & \multicolumn{2}{c}{\begin{tabular}[c]{@{}c@{}}0.4986\\ (0.4711)\end{tabular}}            & \multicolumn{2}{l}{\begin{tabular}[c]{@{}l@{}}0.4833\\ (0.4729)\end{tabular}}           \\ \cline{2-14} 
                                                                      & \multicolumn{1}{c|}{\begin{tabular}[c]{@{}c@{}}Amnesiac\\ Unlearning\end{tabular}}    & \multicolumn{2}{c}{\begin{tabular}[c]{@{}c@{}}0.5004\\ (0.5001)\end{tabular}}            & \multicolumn{2}{c|}{\begin{tabular}[c]{@{}c@{}}0.4987\\ (0.4878)\end{tabular}}           & \multicolumn{2}{c}{\begin{tabular}[c]{@{}c@{}}0.4893\\ (0.4801)\end{tabular}}            & \multicolumn{2}{c|}{\begin{tabular}[c]{@{}c@{}}0.4707\\ (0.4705)\end{tabular}}           & \multicolumn{2}{c}{\begin{tabular}[c]{@{}c@{}}0.4911\\ (0.4920)\end{tabular}}            & \multicolumn{2}{l}{\begin{tabular}[c]{@{}l@{}}0.4954\\ (0.4905)\end{tabular}}           \\ \cline{2-14} 
                                                                      & \multicolumn{1}{c|}{Mini-Unlearning}                                                  & \multicolumn{2}{c}{\textbf{\begin{tabular}[c]{@{}c@{}}0.4453\\ (0.6031)\end{tabular}}}   & \multicolumn{2}{c|}{\textbf{\begin{tabular}[c]{@{}c@{}}0.6094\\ (0.7108)\end{tabular}}}  & \multicolumn{2}{c}{\textbf{\begin{tabular}[c]{@{}c@{}}0.3843\\ (0.4125)\end{tabular}}}   & \multicolumn{2}{c|}{\textbf{\begin{tabular}[c]{@{}c@{}}0.6064\\ (0.6798)\end{tabular}}}  & \multicolumn{2}{c}{\textbf{\begin{tabular}[c]{@{}c@{}}0.4217\\ (0.5546)\end{tabular}}}   & \multicolumn{2}{l}{\textbf{\begin{tabular}[c]{@{}l@{}}0.5412\\ (0.6113)\end{tabular}}}  \\ \bottomrule[1pt]
\end{tabular}
\label{table2}
\end{table}

\paragraph{Test Accuracy of Unlearned Models.} The experimental results are illustrated in Table \ref{table1}. A discernible trend is observed where test accuracy diminishes with an increase in the unlearning ratio, $\eta$, irrespective of the unlearning method employed. This decrement is consistent with expectations. An increased $\eta$ signifies a greater volume of data being excluded from the training process, effectively simulating a scenario where the model is trained with a reduced dataset size. Furthermore, a comparative analysis reveals a notable superiority of Mini-Unlearning, which is the method that mimics retraining followed by DeltaGrad. Certified Data Removal and Amnesiac Unlearning, where the removed samples are limited to $10^2$ samples in original papers, encounter severe failures under high unlearning ratios.

\paragraph{Defensive Ability over MIA.} In the assessment of models' resilience to security breaches post-unlearning, we adopt the methodology delineated in \cite{b27} for training a binary classifier, $\mathcal{A}$. This classifier is designed to ascertain the membership status of individual data samples. The approach in \cite{b27} capitalizes on the differential responses of the attacked model when processing member versus non-member data. Upon the development of $\mathcal{A}$, each data sample $d_i$ or $d_j$ is processed through the post-unlearning models to generate output. This output is then fed into $\mathcal{A}$ to determine the membership status. The precision and recall metrics obtained from this analysis are documented in Tables ~\ref{table2}. 

Certified Data Removal and Amnesiac Unlearning, which fail unlearning, exhibit random guessing over unlearned samples and retained samples. For the comparison of retraining, DeltaGrad, and Mini-Unlearning, Mini-Unlearning achieves the lowest precision/recall over unlearned samples, which means the effectiveness of erasing private information. However, those three methods seem to still remember the private information about the retained samples where MIA could still infer membership information(precision and recall are all greater than 0.5).

\section{Conclusion and Future Work}
This study introduced Mini-Unlearning, a pioneering method in the field of machine unlearning, designed to efficiently remove data traces while enhancing both the accuracy and the resistance of models to membership inference attacks. Mini-Unlearning utilizes a minimal subset of historical gradients and exploits contraction mapping to deal with higher unlearning ratios. The empirical results validate our theoretical claims, demonstrating that Mini-Unlearning outperforms traditional unlearning techniques in terms of accuracy and security.

Looking ahead, we aim to further refine Mini-Unlearning by optimizing the selection of gradients used in the unlearning process, which is expected to enhance the method's efficiency and effectiveness. Additionally, we plan to extend our evaluations to include a wider range of privacy attacks and assess the robustness of Mini-Unlearning against these varied threats.

% \section{Boarder Impacts}
% This paper bridges the theoretical underpinnings and practical applications of an innovative machine unlearning technique, Mini-Unlearning. Our findings offer a deeper understanding of how contraction mapping and minimal gradient dependence can enhance the effectiveness and scalability of machine unlearning processes. This contributes to practical strategies for improving data deletion without sacrificing model integrity or performance. As the use of machine learning with sensitive data proliferates, ensuring that data can be reliably and securely forgotten becomes imperative. Our work furnishes critical insights for developers and practitioners, empowering them to implement robust, privacy-preserving machine learning applications. 

%\authorcontribution{Tao Huang led the whole project, came up with the initial idea, and implemented the algorithm. Guolong Zheng and Jianshan Zhang conducted the experiment. Xuechao Yang collected training data and verified the experiment results. Jingda Yu and Ziyang Chen analyzed the state of the art and suggested improvements. Xu Yang helped refine the algorithm and provided suggestions on experiments and writing. The authors read and approved the final manuscript.}

%\funding{}

%\dataavailability{Data in this paper is available from the corresponding authors upon request.} 

%\conflictsofinterest{The authors declare no conflict of interest.} 

\bibliographystyle{named}
\bibliography{neurips_2024}

\appendix

% \section{Appendix / supplemental material}

\section{Proof of Theorem 1}\label{Appendix A}

\begin{proof} [Proof of Theorem 1]
We start with the general cases. A significant challenge arises in determining the quantity of data records to be sampled from the unlearned dataset for mini-batch Stochastic Gradient Descent (SGD). Let $\mathbf{w}_s$ and $\mathbf{w}_s^{*}$ be the parameters evaluated on the full training set and on the retained set, respectively. Then we have:
\begin{small}
\begin{equation}
    \begin{split}
       \mathbf{w}_s 
       &= \mathbf{w}_{s-1} - \frac{\eta }{B} \sum_{j\in \mathbb{B}_{s}} \nabla F^{(s-1,j)}\left ( \mathbf{w}_{s-1} \right ) \\
       &= \mathbf{w}_{s-1} - \frac{\eta }{B} \sum_{j\in \mathbb{B}_{s}/\bar{\mathbb{B}}_{s}} \nabla F^{(s-1,j)}\left ( \mathbf{w}_{s-1} \right ) - \frac{\eta }{B} \sum_{j\in \bar{\mathbb{B}}_{s}} \nabla F^{(s-1,j)}\left ( \mathbf{w}_{s-1} \right ),\\
       \mathbf{w}_s^{*}&= \mathbf{w}_{s-1}^{*} - \frac{\eta }{B-\triangle B} \sum_{j\in \mathbb{B}_{s}/\bar{\mathbb{B}}_{s}} \nabla F^{(s-1,j)}\left ( \mathbf{w}_{s-1}^{*} \right )
    \end{split}
    \label{s-eq-1}
\end{equation}
\end{small}

% \begin{small}
% \begin{equation}
%     \begin{split}
%         \mathbf{w}_s^{*}= \mathbf{w}_{s-1}^{*} - \frac{\eta }{B-\triangle B} \sum_{j\in \mathbb{B}_{s}/\bar{\mathbb{B}}_{s}} \nabla F^{(s-1,j)}\left ( \mathbf{w}_{s-1}^{*} \right )
%     \end{split}
%     \label{s-eq-2}
% \end{equation}
% \end{small}

The discrepancy, denoted as $\Delta \mathbf{w}_s$, between the parameter values derived from the complete training set and those obtained from the retained set is articulated in Eq.(\ref{s-eq-3}). This equation embodies an inductive approach. Specifically, $\Delta \mathbf{w}_{s-1}$ can be substituted by $\Delta \mathbf{w}_{s-2}$, where $\Delta \mathbf{w}_{s-1}$ is represented using $\Delta \mathbf{w}_{s-2}$, with a modification in the subscript from $s$ to $s-1$. The computation of $\Delta \mathbf{w}_s$ for the initial epoch is straightforward, enabling the determination of $\Delta \mathbf{w}_s$ for subsequent epochs.
\begin{small}
\begin{equation}
    \begin{split}
        &\Delta \mathbf{w}_s= \mathbf{w}_{s}^{*} -\mathbf{w}_s = \Delta \mathbf{w}_{s-1} + \frac{\eta }{B} \sum_{j\in \mathbb{B}_{s}/\bar{\mathbb{B}}_{s}} \nabla F^{(s-1,j)}\left ( \mathbf{w}_{s-1} \right ) \\
        &\quad - \frac{\eta }{B-\triangle B} \sum_{j\in \mathbb{B}_{s}/\bar{\mathbb{B}}_{s}} \nabla F^{(s-1,j)}\left ( \mathbf{w}_{s-1}^{*} \right )  + \frac{\eta }{B} \sum_{j\in \bar{\mathbb{B}}_{s}} \nabla F^{(s-1,j)}\left ( \mathbf{w}_{s-1} \right )\\
        &=\Delta \mathbf{w}_{s-1} + \frac{\eta }{B} \sum_{j\in \mathbb{B}_{s}/\bar{\mathbb{B}}_{s}} \nabla F^{(s-1,j)}\left ( \mathbf{w}_{s-1} \right ) + \frac{\eta }{B} \sum_{j\in \bar{\mathbb{B}}_{s}} \nabla F^{(s-1,j)}\left ( \mathbf{w}_{s-1} \right )\\
        &\quad - \frac{\eta }{B-\triangle B} \sum_{j\in \mathbb{B}_{s}/\bar{\mathbb{B}}_{s}} \left ( \nabla F^{(s-1,j)}\left ( \mathbf{w}_{s-1} \right ) + \nabla^{2} F^{(s-1,j)}\left ( \mathbf{w}_{s-1} \right ) \cdot \Delta \mathbf{w}_{s-1} \right ) \\
        &=\frac{\eta }{B} \left ( \frac{\gamma_s}{1+\gamma_s}  \right )  \sum_{j\in \mathbb{B}_{s}/\bar{\mathbb{B}}_{s}} \nabla F^{(s-1,j)}\left ( \mathbf{w}_{s-1} \right )  -\frac{\eta }{B} \sum_{j\in \bar{\mathbb{B}}_{s}} \nabla F^{(s-1,j)}\left ( \mathbf{w}_{s-1} \right ) \\
        &\quad + \left ( I - \frac{\eta }{B-\triangle B} \sum_{j\in \mathbb{B}_{s}/\bar{\mathbb{B}}_{s}} \nabla^{2} F^{(s-1,j)}\left ( \mathbf{w}_{s-1} \right )  \right ) \cdot \Delta \mathbf{w}_{s-1}
    \end{split}
    \label{s-eq-3}
\end{equation}
\end{small}

For the sake of simplicity, we define $G\left ( \mathbb{B}_{l}, \bar{\mathbb{B}}_{l}, \mathbf{w}_{l-1}   \right )$ and $H\left ( \mathbb{B}_{l}, \bar{\mathbb{B}}_{l}, \mathbf{w}_{l-1}   \right )$ as follows:
\begin{small}
\begin{equation}
\begin{split}
&G\left ( \mathbb{B}_{l}, \bar{\mathbb{B}}_{l}, \mathbf{w}_{l-1}   \right ):=\frac{\eta}{B} \left ( \frac{\gamma _{l}}{1+\gamma _{l}}  \sum_{j \in \mathbb{B}_{l}/\bar{\mathbb{B}}_{l}} \nabla F^{(l-1,j)}\left ( \mathbf{w}_{l-1} \right )+\sum_{j \in \bar{\mathbb{B}}_{l}} \nabla F^{(l-1,j)}\left ( \mathbf{w}_{l-1} \right ) \right)\\
&\quad =\frac{\eta}{B} \left( \frac{B}{B-\bigtriangleup B_l}  \sum_{j \in \bar{\mathbb{B}}_{l}} \nabla F^{(l-1,j)}\left ( \mathbf{w}_{l-1} \right )\right) -\frac{\eta}{B} \left(\frac{\bigtriangleup B_l}{B-\bigtriangleup B_l}   \sum_{j \in \mathbb{B}_{l}} \nabla F^{(l-1,j)}\left ( \mathbf{w}_{l-1} \right) \right).
\end{split}
\label{s-eq-4}
\end{equation}
\end{small}

\begin{small}
\begin{equation}
\begin{split}
&H\left ( \mathbb{B}_{l}, \bar{\mathbb{B}}_{l}, \mathbf{w}_{l-1}   \right ) = I - \frac{\eta}{B-\bigtriangleup B_{l}} \sum_{j \in \mathbb{B}_{l} /\bar{\mathbb{B}}_{l}} \nabla^{2} F^{(l-1,j)}\left ( \mathbf{w}_{l-1} \right ) 
\end{split}
\label{s-eq-5}
\end{equation}
\end{small}

Substituting Eq.(\ref{s-eq-4}) and Eq.(\ref{s-eq-5}) into Eq.(\ref{s-eq-3}) and pushing forward $k$ rounds, we can get:
\begin{small}
\begin{equation}
    \begin{split}
        &\Delta \mathbf{w}_s = G\left ( \mathbb{B}_s,\bar{\mathbb{B} }_s, \mathbf{w}_{s-1}   \right ) +H\left ( \mathbb{B}_s,\bar{\mathbb{B} }_s, \mathbf{w}_{s-1}   \right ) \cdot \Delta \mathbf{w}_{s-1}\\
        &= G\left ( \mathbb{B}_s,\bar{\mathbb{B} }_s, \mathbf{w}_{s-1}   \right ) +H\left ( \mathbb{B}_s,\bar{\mathbb{B} }_s, \mathbf{w}_{s-1}   \right ) \cdot G\left ( \mathbb{B}_{s-1},\bar{\mathbb{B} }_{s-1}, \mathbf{w}_{s-2}   \right )\\
        &\quad + H\left ( \mathbb{B}_s,\bar{\mathbb{B} }_s, \mathbf{w}_{s-1}   \right )\cdot H\left ( \mathbb{B}_{s-1},\bar{\mathbb{B} }_{s-1}, \mathbf{w}_{s-2}   \right ) \cdot  \Delta \mathbf{w}_{s-2} = \cdots \\
        &= G\left ( \mathbb{B}_s,\bar{\mathbb{B} }_s, \mathbf{w}_{s-1}   \right ) + \sum_{j=2}^{k} \left ( \prod_{i=1}^{j-1} H\left ( \mathbb{B}_{s-i+1},\bar{\mathbb{B} }_{s-i+1}, \mathbf{w}_{s-i}   \right ) \right ) \cdot G\left ( \mathbb{B}_{s-j+1},\bar{\mathbb{B} }_{s-j+1}, \mathbf{w}_{s-j}   \right )\\
        &\quad +\left ( \prod_{i=1}^{k} H\left ( \mathbb{B}_{s-i+1},\bar{\mathbb{B} }_{s-i+1}, \mathbf{w}_{s-i}   \right ) \right ) \cdot \Delta \mathbf{w}_{s-k} 
    \end{split}
    \label{s-eq-6}
\end{equation}
\end{small}

So the following inequality holds:
\begin{small}
\begin{equation}
    \begin{split}
        &\left \| \Delta \mathbf{w}_s -\left ( G\left ( \mathbb{B}_s,\bar{\mathbb{B} }_s, \mathbf{w}_{s-1}   \right ) + \sum_{j=2}^{k} \left ( \prod_{i=1}^{j-1} H\left ( \mathbb{B}_{s-i+1},\bar{\mathbb{B} }_{s-i+1}, \mathbf{w}_{s-i}   \right ) \right ) \cdot G\left ( \mathbb{B}_{s-j+1},\bar{\mathbb{B} }_{s-j+1}, \mathbf{w}_{s-j}   \right ) \right )  \right \|_2\\
        &\quad \le \left \| \left ( \prod_{i=1}^{k} H\left ( \mathbb{B}_{s-i+1},\bar{\mathbb{B} }_{s-i+1}, \mathbf{w}_{s-i}   \right ) \right ) \cdot \Delta \mathbf{w}_{s-k} \right \|_2 \le \left \| \left ( \prod_{i=1}^{k} H\left ( \mathbb{B}_{s-i+1},\bar{\mathbb{B} }_{s-i+1}, \mathbf{w}_{s-i}   \right ) \right ) \right \|_2 \cdot \left \| \Delta \mathbf{w}_{s-k} \right \|_2 \\
        &\quad \le  r^k \left \| \Delta \mathbf{w}_{s-k} \right \|_2
    \end{split}
    \label{s-eq-7}
\end{equation}
\end{small}
where $r = \text{max}\{ \|1-\eta \cdot  \mu \|,  \|1-\eta \cdot  L \|\} \in (0,1)$.

The first and second inequalities are the properties of the matrix norm. The last inequality holds since the eigenvalues of $H\left ( \mathbb{B}_{s-i+1},\bar{\mathbb{B} }_{s-i+1}, \mathbf{w}_{s-i}   \right )$ are smaller than $r$.
\end{proof}

\section{ How to Compute $H\left ( \mathbb{B}_{k+1},\bar{\mathbb{B} }_{k+1}, \mathbf{w}_{k}   \right ) \cdot \mathbf{u}$}\label{Appendix B}

\begin{algorithm} [tb] 
	\caption{Algorithm to calculate the product $H\left ( \mathbb{B}_{k+1},\bar{\mathbb{B} }_{k+1}, \mathbf{w}_{k}   \right ) \cdot \mathbf{u}$} 
	\label{alg4} 
            \KwIn{the gradients $\{ \nabla F\left ( \mathbf{w}_{k-m} \right ),\cdots,\nabla F\left ( \mathbf{w}_{k} \right ) \}$, the model parameters $\{ \mathbf{w}_{k-m},\cdots,\mathbf{w}_k \}$, a vector $\mathbf{u}$, history size $m$.}
            \KwOut{Approximate results of $H\left ( \mathbb{B}_{k+1},\bar{\mathbb{B} }_{k+1}, \mathbf{w}_{k}   \right ) \cdot \mathbf{u}$.}
            Compute $\Delta W = \{ \Delta \mathbf{w}_0, \Delta \mathbf{w}_1, \cdots, \Delta \mathbf{w}_{m-1} \}$ such that $\Delta \mathbf{w}_i = \mathbf{w}_{k-m+i+1} - \mathbf{w}_{k-m+i}$; \\
            Compute $\Delta G = \{ \Delta \mathbf{g}_0, \Delta \mathbf{g}_1, \cdots, \Delta \mathbf{g}_{m-1} \}$ such that $\Delta \mathbf{g}_i = \nabla F\left ( \mathbf{w}_{k-m+i+1} \right ) - \nabla F\left ( \mathbf{w}_{k-m+i} \right )$;\\
            Compute $\Delta W^T \Delta W$;\\
            Compute $\Delta W^T \Delta G$, get its diagonal matrix $D$ and its lower triangular submatrix $L$;\\
            Compute $\sigma = \frac{\Delta \mathbf{g}_{m-1}^T \Delta \mathbf{w}_{m-1}}{\mathbf{w}_{m-1}^T \mathbf{w}_{m-1}}$;\\
            Compute the Cholesky factorization for $\sigma \Delta W^T \Delta W + LDL^T$ to get $JJ^T$;\\
            Compute: $p=\left [ \begin{matrix}
 - D^{\frac{1}{2} } & - D^{\frac{1}{2}} L^T \\ 
 \mathbf{0}  & J^T
\end{matrix} \right ]^{-1} \left [ \begin{matrix}
 - D^{\frac{1}{2} } & \mathbf{0} \\  
 - D^{\frac{1}{2}} L^T  & J^T
\end{matrix} \right ]^{-1}\left [ \begin{matrix}
\Delta G^T \mathbf{u}  \\ \sigma \Delta W^T \mathbf{u}
\end{matrix} \right ] $; \\
           Compute $\Bar{H}\left ( \mathbb{B}_{k+1},\bar{\mathbb{B} }_{k+1}, \mathbf{w}_{k}   \right ) \cdot \mathbf{u} = \sigma \mathbf{u} - \left [ \begin{matrix}
 \Delta G & \sigma \Delta W
\end{matrix} \right ] p$;\\
           \Return{$H\left ( \mathbb{B}_{k+1},\bar{\mathbb{B} }_{k+1}, \mathbf{w}_{k}   \right ) \cdot \mathbf{u} =I \cdot \mathbf{u} - \frac{\eta}{B-\bigtriangleup B_{k+1}} \Bar{H}\left ( \mathbb{B}_{k+1},\bar{\mathbb{B} }_{k+1}, \mathbf{w}_{k}   \right ) \cdot \mathbf{u}$}
\end{algorithm}

The Limited-memory Broyden-Fletcher-Goldfarb-Shanno algorithm, commonly known as L-BFGS, is an optimization algorithm in the family of quasi-Newton methods. It is particularly well-suited for solving large-scale optimization problems where the objective function is smooth, but potentially complex and high-dimensional. The efficiency of L-BFGS lies in its ingenious approach to approximating the Hessian matrix, which is crucial for Newton's method in optimization. The true efficiency of L-BFGS becomes apparent when \textbf{\textit{calculating the product of a matrix and a vector}}, a common operation in optimization algorithms. In standard methods, this calculation involves the actual Hessian matrix and can be computationally intensive. L-BFGS, however, utilizes its approximate inverse Hessian. This approximation significantly reduces the computational complexity and memory requirements, as it avoids the direct calculation and storage of the full Hessian matrix.

In our cases, let $\bar{H}\left ( \mathbb{B}_{k+1},\bar{\mathbb{B} }_{k+1}, \mathbf{w}_{k}   \right ) = \sum_{j \in \mathbb{B}_{k+1} /\bar{\mathbb{B}}_{k+1}} \nabla^{2} F^{(k,j)}\left ( \mathbf{w}_{k} \right )$ where is a Hessian matrix, the Hessian-vector product $\bar{H}\left ( \mathbb{B}_{k+1},\bar{\mathbb{B} }_{k+1}, \mathbf{w}_{k}   \right ) \cdot \mathbf{u}$ could be calculated via L-BFGS algorithm. The hyper-parameter $m$ in L-BFGS determines the number of the components in the lists of parameter differences $\Delta W$ and gradient differences $\Delta G$. Larger $m$ contributes to better approximation. In our experiments, we choose $m=2$. The following algorithm presents the process to calculate $H\left ( \mathbb{B}_{k+1},\bar{\mathbb{B} }_{k+1}, \mathbf{w}_{k}   \right ) \cdot \mathbf{u}$. In Algorithm \ref{alg4}, $\nabla F\left ( \mathbf{w}_{k} \right )$ represents $\sum_{j\in \mathbb{B}_{k+1}/\bar{\mathbb{B}}_{k+1}} \nabla F^{(k,j)}\left ( \mathbf{w}_{k} \right )$.

\section{Ablation Study}
Table \ref{abla} shows test accuracy over different datasets with various $k$ under $\eta = 5\%$. The results indicate that the larger the value of $k$, the higher the accuracy of the model on the test set after unlearning. Additionally, the impact of the $k$ value on accuracy is not very significant. If a more efficient unlearning process is needed and the requirement for test accuracy is not as high, a smaller $k$ value can be chosen to achieve data unlearning.

\begin{table}[htb]
\caption{Ablations for $k$}
\centering
\begin{tabular}{c|ccccc}
\toprule[1pt]
$k$       & 2    & 4    & 6    & 8    & 10   \\ \midrule
MNIST   & 0.80 & 0.81 & 0.81 & 0.82 & 0.82 \\ 
Covtyoe & 0.68 & 0.68 & 0.69 & 0.69 & 0.70 \\ 
HIGGS   & 0.61 & 0.61 & 0.62 & 0.64 & 0.64 \\ \bottomrule[1pt]
\end{tabular}
\label{abla}
\end{table}

% %% The file named.bst is a bibliography style file for BibTeX 0.99c

\end{document}